\documentclass[preprint,review,12pt]{elsarticle}
\usepackage{amsmath}
\usepackage{hyperref}
\usepackage[labelfont=bf]{caption}
\usepackage{multirow}
\journal{International Journal of Computational Intelligence Systems}
\begin{document}
\begin{frontmatter}

\title{Enhancing Multi-Step Brent Oil Price Forecasting with Ensemble Multi-Scenario Bi-GRU Networks}

\author[uinvr] {Mohammed Alruqimi}
\ead{mohammed.alruqimi@univr.it}
\cortext[cor1]{corresponding author}

\author[uinvr]{Luca Di Persio}
\ead{luca.dipersio@univr.it}

\affiliation[uinvr]{organization={University of Verona},%Department and Organization
            %addressline={}, 
            city={Verona},
            %postcode={}, 
            %state={},
            country={Italy}}

\begin{abstract}
Despite numerous research efforts in applying deep learning to time series forecasting, achieving high accuracy in multi-step predictions for volatile time series like crude oil prices remains a significant challenge. Moreover, most existing approaches primarily focus on one-step forecasting, and the performance often varies depending on the dataset and specific case study.
 In this paper, we introduce an ensemble model to capture Brent oil price volatility and enhance the multi-step prediction.
 Our methodology employs a two-pronged approach. First, we assess popular deep-learning models and the impact of various external factors on forecasting accuracy. Then, we introduce an ensemble multi-step forecasting model for Brent oil prices. Our approach generates accurate forecasts by employing ensemble techniques across multiple forecasting scenarios using three BI-GRU networks.
 Extensive experiments were conducted on a dataset encompassing the COVID-19 pandemic period, which had a significant impact on energy markets. The proposed model's performance was evaluated using the standard evaluation metrics of MAE, MSE, and RMSE. The results demonstrate that the proposed model outperforms benchmark and established models.

\end{abstract}

%%Graphical abstract
%\begin{graphicalabstract}
%\includegraphics{grabs}
%\end{graphicalabstract}

%%Research highlights
%\begin{highlights}
%\item An ensemble model for multi-step Brent crude oil price forecasting.
%\item Incorporating an accumulative sentiment index and residual component features through two bidirectional GRU models. 
%\item Evaluated the influence of integrating various external factors to improve prediction performance.

%\item GRU and LSTM networks exhibit superior predictive performance compared to Transformer-based models.

%\item Modeling price volatility based on sentiment index and time series decomposition

%\end{highlights}

\begin{keyword}
%% keywords here, in the form: keyword \sep keyword
Crude oil price forecasting \sep Brent oil forecasting \sep sentiment Bi-GRU \sep time series forecasting  

%% PACS codes here, in the form: \PACS code \sep code

%% MSC codes here, in the form: \MSC code \sep code
%% or \MSC[2008] code \sep code (2000 is the default)

\end{keyword}

\end{frontmatter}

%% \linenumbers

%% main text
\section{Introduction}
\label{sec:introduction}
Accurate predictions of oil prices are essential due to their vital role in the global economy. However, the prediction of crude oil prices is renowned for its obscurity and complexity. The high degree of volatility, unpredictable, irregular events, and complex interconnections among market factors make it extremely challenging to accurately forecast the fluctuations in crude oil prices.  The dynamic interplay of supply and demand and changes in oil prices are influenced by external factors such as economic growth, financial markets, geopolitical conflicts, warfare, and political considerations \cite{inf.factors, Lu2021,JIANG2022123471}. 
   A variety of methodologies have been utilised for predicting crude oil prices, involving the application of econometric and statistical time series analysis techniques such as VAR \cite{DRACHAL2021102244}, ARIMA, GARCH \cite{MOHAMMADI20101001}, VMD \cite{HUANG2021106669}, and Walvet decomposition \cite{HE2012564}.  In more recent studies, there has been a prevalent use of machine learning models and hybrid approaches \cite{Lu2021,article123,LINDEMANN2021650} in the literature.  Nevertheless, achieving accurate oil price forecasting remains a challenging task, particularly in terms of multi-step forecasting.  Traditional econometric and statistical methods are often inadequate for forecasting oil prices due to many challenges related to the irregular characteristics of energy markets, such as non-stationarity, multi-frequency, non-linearity, and chaotic properties \cite{RePEc-ftiti}.  To overcome these challenges, AI models have gained significant popularity for oil price prediction.  In contrast to traditional statistical and econometric models, AI provides a valuable alternative approach to capture complex nonlinear characteristics of the crude oil price movement. There is an effort to exploit the emergent advances in deep-learning machine models (such as Transformers and GANs) for time series analysis \cite{abs-2106-13008,Deng2021}.  Despite the great success achieved by these models in processing natural languages and generating images and videos, Long short-term memory (LSTM) and gated recurrent unit (GRU) networks still maintain their popularity for time series forecasting \cite{EREN2024114031,zeng2022transformers, SHEN2023953}.  Their popularity could be attributed to their ability to effectively capture temporal dependencies, which is crucial for accurately predicting future values in a time series.  They are also significantly less complex to train than more modern models such as GANs and transformers.  Also, a vast amount of data is required to train transformers, which makes them inefficient for tasks involving smaller datasets, such as daily-based oil price forecasting.
   Additionally, most current research focuses on one-day-ahead forecasting or utilises complex combinations of layers and models with high computational complexity.  Despite these efforts, multi-step forecasting accuracy remains suboptimal. 
   
   This study proposes to blind forecasts generated from three different scenarios across Bi-GRU networks. 
   To this end, we start by assessing the forecast performance of various architectures of deep learning models( popular deep learning models used in the literature for price forecasting) to optimise our model selection.  Then, we propose a novel effective model (Abbreviated as ERS-Bi-GRU, denoting its main elements—Ensemble, Residual, Sentimental, and Bi-Directional Gated Recurrent Units) for Brent price forecasting.  Our experiments additionally evaluate the influence of incorporating three external factors – the USD index (USDX), Saudi energy sector index (TENI), and sentiment score (SENT) – on enhancing prediction accuracy.  
   The contributions of this paper are threefold:

    \begin{itemize}
        \item Introduction of an effective ensemble model for forecasting multi-step Brent crude oil prices.
        \item Evaluation of various established deep learning architectures and combinations for oil price forecasting, discerning the optimal architecture within this domain. To the best of our knowledge, this is the first study to compare RNN-based models and more recent and complex models like transformers for oil price prediction.
        \item Evaluate the influence of external factors on Brent price movement. 
    \end{itemize}
    The proposed model (ERS-Bi-GRU) has been compared against well-known benchmarks and established models in the literature to evaluate its forecasting accuracy.  The obtained results indicate that the proposed model outperforms the benchmark models. 
    The rest of the paper is organised as follows. 
    Section (\ref{sec:litrature-review}) provides an overview of recent literature related to oil price forecasting.
    Section(\ref{sec:dataset}) describes the dataset used in this study.  Section (\ref{sec:methodology}) outlines the methodology, Section (\ref{sec:experiment}) describes the experimental setup and results, and finally the conclusion.

\section{Literature review}
\label{sec:litrature-review}

        Recent years have witnessed strong growth in adopting machine learning for time series forecasting, driven by its remarkable ability to uncover intricate and non-linear patterns within the data. Researchers have employed various Ml networks such as long-short-term memory (LSTM) \cite{JIANG2022123471,Deng2021, su142114616}, gated recurrent unite (GRU) \cite{WANG2020104827,ZHANG2023119617,FANG2023119329}, convolutional neural network (CNN) \cite{HE2022959}, and transformer models \cite{LIU2023e16715} to predict crude oil prices based on historical price data and some relevant features. Machine learning approaches often require extensive feature engineering and can be sensitive to the quality and availability of data. Many researchers incorporated additional data sources, such as macroeconomic and technical indicators, social media sentiment, and news articles, to enhance prediction accuracy. Additionally, advancements in natural language processing and sentiment analysis techniques have allowed for a more comprehensive understanding of the impact of geopolitical events and news on oil prices \cite{JIANG2022123471,FANG2023119329,Kaplan2023, ZHAO2023103506}.
        A popular approach combines different neural networks with statistical and economic methods to improve crude oil forecasting. For instance, many researchers have merged RNN networks with CNN and self-attention mechanisms to capture temporal, local, and long-term dependencies in historical price data \cite{refId0,SHAN2021108838}. Statistical and time series analysis methods (such as variational mode decomposition (VMD), empirical mode decomposition (EMD), Granger causality, Gaussian process, complete ensemble empirical mode decomposition with adaptive noise (CEEMDAN)) are also prevalent alongside neural networks in recent approaches 
       (i.e. \cite{JIANG2022123471,FANG2023119329}).
        
        %%\begin{table}[!hbt]
          %%  \centering
           %% \hrulefill
            %\caption{A review of recent related works}
            %\label{tab:review}
            %\begin{tabular}{p{3cm} | p{4.5cm} | p{5cm}} 
            %\hline	
            %\end{tabular}
        %\end{table}

\section{Preliminaries}
\label{sec:Preliminaries}
        As listed in Table \ref{tab:expermint_design}, the targeted networks selected for the experiments include LSTM, GRU, CNN-LSTM, CNN-LSTM-att, Transformer, Autoformer, Informer, and TimsNet. GRU and LSTM networks have been selected due to their established effectiveness in time series forecasting tasks \cite{Deng2021, WANG2020104827}. Combining CNN and LSTM networks and attention is another popular approach widely employed in the literature. CNNs excel at extracting local features from time series data by applying learnable filters that capture spatial relationships within specific time windows.The CNN extracts relevant local features, while the LSTM and GRU model long-term dependencies. Additionally, the self-attention mechanisms effectively directs attention across the time series data.
        On the other hand, transformers have demonstrated remarkable achievements across various domains beyond just NLP. Over the last three years, transformer-based architectures (including Informer \cite{2012-07436}, Autoformer \cite{abs-2106-13008}) have been adapted for time series tasks. TimesNet \cite{wu2023timesnet} is another cutting-edge CNN-based-structure model introduced in April (2023).  

\section{Dataset}
\label{sec:dataset}

    \subsection{Dataset description}
    The initial dataset used in this paper encompasses eight variables: closing prices of Brent oil, USD index, Saudi Energy index, Saudi Tadawul All Share index, S\&P 500 index, Natural Gas index, Gold index, and a sentimental score. These variables span 2,380 observations from January 2012, to April 2021, capturing the period impacted by the COVID-19 pandemic on energy and stock markets. This time range was specifically chosen due to the availability of sentiment scores for this period.
        \begin{figure}[!hbt]
                 \centering
                 \includegraphics[scale=0.55]{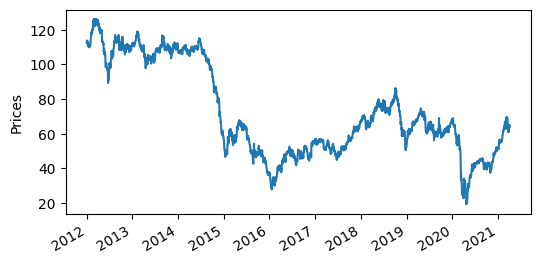}
                 \caption{Brent oil price trend from 2012 to 2021}
                 \label{fig:oil-price}
        \end{figure}
        Figure (\ref{fig:oil-price}) displays the time-series plot of Brent's daily closing price. The figure shows that sharp fluctuations in crude oil logistics and production are often linked to world events. 
    Extensive research has explored the influence of industrial commodities, indices (such as USD, S\&P 500, Natural Gaz, and Gold), and sentiment analysis extracted from news articles, financial reports, and even tweets on the energy market \cite{COLEMAN2012318,MIAO2017776,2017,Kaplan2023,ZHAO2023103506}. 
    External factors that affect the movement of crude oil prices could be categorised into supply, demand, financial, political, and major events factors \cite{COLEMAN2012318}. These factors are inter-correlated; for instance, supply and demand movement may be subject to political factors and world events.
    
    \subsection{Correlation-based Feature Selection}
\begin{figure}[!hbt]
                \centering
                \includegraphics[scale=0.7]{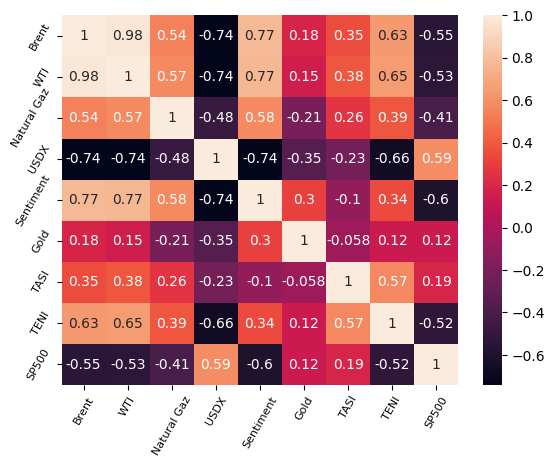}
                \caption{Heatmap Spearman Correlation}
                \label{fig:Speraman}
    \end{figure}

    Correlation-based Feature Selection is a method to identify and select the most relevant features from a dataset by analysing the relationship between each feature and the target variable. This technique helps select features that can play the most significant role in predicting the target variable.
    We created a filtered dataset includes only three external variables from our initial dataset, to be used as input features to our models alongside the Brent oil prices, based on their strong correlation with Brent prices. We used a Spearman correlation coefficient threshold of 0.6 to identify these variables, as shown in Figure (\ref{fig:Speraman}).
    Therefore, the three variables enumerated below have been designated for the subsequent experimentation alongside the target variable (Brent crude oil prices). The impact of these variables on the prediction performance has been further evaluated experimentally during the models training, as detailed in Section \ref{sec:experiment} 

        \begin{enumerate}
             \item Sentiment score (SENT): We used an accumulative sentimental score provided by CrudeBERT\_Plus model and presented in \cite{Kaplan2023}. CrudeBERT is a variant of FinBERT that has been fine-tuned towards assessing the impact of market events on crude oil prices, focusing on frequently occurring market events and their effects on market prices according to Adam Smith’s theory of supply and demand. Mainly, CrudeBERT dataset used headlines originating from 1034 unique news sources, of which the majority has been published on the Dow Jones newswires (approx. 21,200), followed by Reuters (approx. 3,000), Bloomberg (approx. 1,100), and Platts (approx. 870). More details about generating this sentiment score are found in the paper \cite{Kaplan2023}; the data is publicly available and can be reached following this \href{https://github.com/fhgr/crudeoil-sentiment2022-dataset}{link}. 
             \item USDX Index: The U.S. dollar index (USDX) measures the value of the U.S. dollar relative to a basket of foreign currencies. There is a negative correlation between crude oil and the USDX index. USDX historical dataset has been obtained for the same period from \href{https://www.investing.com}{Investing}. 
             \item Saudi energy sector index (TENI) contains two companies working in the energy sector (Arabian Drilling Co and Rabigh Refining \& Petrochemical Co). The historical dataset of this index has been obtained for the same period from \href{Investing} {https://uk.investing.com/indices/tpisi-historical-data}.
        \end{enumerate}
        
    \subsection{Data preparation}
        Our training data have been standardised by removing the mean and scaling to unit variance. To incorporate observations of all features, we run a left-join merge between the Brent crude oil price time series and the other three input time series. On the other hand, the missing values in the three external factors have been filled in by running a Linear Interpolation function:  
            Let's $X_t$ is a null value, $X_t=($$X_{t-1}+$$X_{t+1})/2$. The dataset has been split into train/valid and test as follows:
            Training set:
            from 2012-01-03 to 2019-10-10; 
            Validation set:
             from 2019-10-11 to 2020-06-23; and
            Test set
 from 2020-06-24 to 2021-04-01.
    
\section{Methodology}
\label{sec:methodology}
    This work presents a hybrid approach, based on deep learning models and times-series volatility analysis, for multi-step forecasting of Brent crude oil prices. Furthermore, it examines the impact of external factors on prediction accuracy, identifies the most suitable deep learning architecture, and optimises time steps and hyperparameters for improved forecasting performance.

\begin{figure}
        \centering
        \includegraphics[scale=0.64]{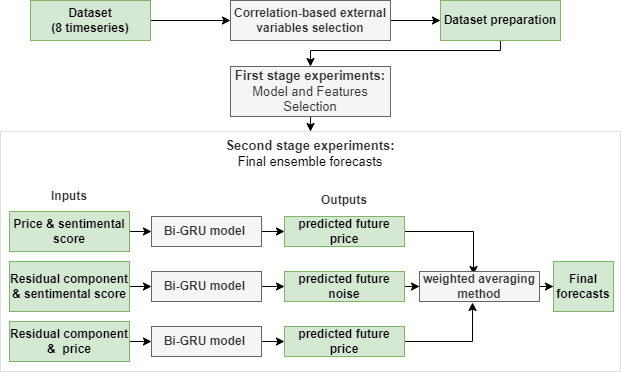}
        \caption{Methodology flowchart}
        \label{fig:ers-methodology}
    \end{figure}

    To this end, we followed the steps outlined below, illustrated by  Figure (\ref{fig:ers-methodology}).
    Firstly, we collected eight historical data (eight-time series), performed a correlation-based test to select only the most correlated variables for the next phase, and prepared our filtered dataset, as described in Section (\ref{sec:dataset}). The filtered dataset comprises daily historical observations of Brent crude oil prices, along with USDX, SENT, and TENI.
    
    Secondly (in the first stage experiments), we conducted experiments using different architectures of GRU, LSTM, CNN and transformer-based models adapted for time series forecasting. A wide range of experiments was carried out, considering various model architectures, time steps and forecasting horizons. We trained each model with univariate and multivariate inputs, incorporating the three external factors (USDX, NETI, and SENT) individually and collectively. We compared the performance of each model in one-step and multi-step horizon forecasting, as discussed in Section (\ref{sec:experiment}). We evaluated the results obtained by each model using MAE, MSE and RMSE metrics. This phase's objective is to identify the most effective model experimentally. It features for our case study, which will be used in the following (final) phase to produce the ensemble forecast.
    
    Finally, based on the results and observations from the previous phase, we built the ERS-Bi-GRU model, which merges forecasts of three Bi-GRU networks performing three forecasting scenarios,  as detailed in Section (\ref{sec:ers-bigru})
    The ERS-GRU model demonstrated superior performance compared to other targeted and created benchmark models.

\section{Experiments}
\label{sec:experiment}
     This section highlights a representative sample of the experiments carried out, the evaluation metrics used, and the best results achieved by each model.
    \subsection{Error evaluation metrics}
        To evaluate the performance of each model, Three commonly used evaluation metrics were employed: Mean Squared Error (MSE), Mean Absolute Error (MAE), and Root Mean Squared Error (RMSE). Mean Absolute Error (MAE) measures the average absolute difference between the predicted and actual values. It is calculated as $$\text{MAE}(y, \hat{y}) = \frac{ \sum_{i=0}^{N - 1} |y_i - \hat{y}_i| }{N}$$
        Mean Squared Error (MSE) calculates the average squared difference between the predicted and actual values. It is computed as $$\text{MSE}(y, \hat{y}) = \frac{\sum_{i=0}^{N - 1} (y_i - \hat{y}_i)^2}{N}$$
        Root Mean Squared Error (RMSE) is the square root of the MSE, providing a measure of the average magnitude of the error. It is given by:
        $$\text{RMSE}(y,\hat{y}) = \sqrt{\frac{\sum_{i=0}^{N - 1} (y_i - \hat{y}_i)^2}{N}}$$
        These evaluation metrics allow for quantifying the accuracy and precision of the model's predictions. Lower MSE, MAE, and RMSE values indicate better performance, indicating smaller discrepancies between the predicted and actual values.
        For a fair comparison, the models have been trained under the same procedures but with a customised hyperparameter configuration for each model. MAE, MSE, and RMSE are calculated using real, not scaled prices. 
    \subsection{Experimental design}  
        Table (\ref{tab:expermint_design}) shows the experiment parameters. To evaluate the performance of each one of the targeted models (GRU, LSTM, Bi-GRU, Bi-LSTM, LSTM-CNN, LSTM-CNN-attention, Autoformer, Informer, Transformer, TimesNet), a systematic approach was employed, considering various window sizes (tuning range from 5  to 22) with one-step(one-ahead) and multi-step (three ahead days) forecasts, with more focus on the multi-step forecasting. For each model, two training scenarios were considered: (1) utilising only Brent crude price lags (univariate) and (2) incorporating different combinations of the three external factors, namely USDX, SENT, and TENI (multivariate). This approach enabled a comprehensive assessment of the models from multiple perspectives. To ensure a unified, systematic comparison, the models were applied to the same dataset as described in Section (\ref{sec:dataset}). A unified dataset splitting and evaluation methodology was adopted to maintain consistency throughout the experiments. Hyperparameters were tuned separately for each model based on the MSE during the training phase, considering various factors such as window size, prediction horizon, input types (univariate/multivariate), and the inclusion of external factors. 
     
        \begin{table}[!hbt]
            \setlength{\tabcolsep}{1pc}
             \hrule
             \caption{Experiments design}
            \label{tab:expermint_design}

            \begin{tabular}{@{}p{4cm}|p{8cm}} \hline 
             Models& LSTM, GRU, Bi-LSTM, Bi-GRU, CNN-Bi-LSTM, CNN-Bi-LSTM-att, Transformer, Autoformer, Informer, TimesNet\\ 
             External variables& USDX, TENI, SENT\\ 
             Window size & tuning range 5 - 22\\ 
             Prediction horizon& 1,3\\ \hline
        \end{tabular} 
        \vspace{3mm}
        \hrule
        \centering
    \end{table}
\section{Findings and Discussion}

     \subsection{LSTM and GRU models}
         Despite the emergence of newer advanced models in the field of deep learning, LSTM and GRU models have maintained their significance and remain competitive options for time series prediction tasks. These models excel in preserving the temporal order. The simplicity of model tuning and their lower computation complexity compared to Transformers or GANs are further advantages. Results obtained from  GRU and LSTM models are shown in Tables (\ref{tab:gru} and \ref{tab:lstm}). The best results achieved were by the SENT-GRU-1 model (MAE 0.0647,	MSE 0.0069, and RMSE 0.0833) in terms of one day ahead and (MAE 1.0411, MSE 2.0097 , and RMSE 0.4176) in term of a 3-days ahead forecasting with the SENT-Bi-GRU model. It is worth mentioning that the evaluation metrics ( MAE, MSE and RMSE) have been calculated using the original values  (re-normalised values). 
     \begin{table}[!hbt]
     \setlength{\tabcolsep}{1.45pc}
        \hrule
        \caption{GRU-based models}
        \label{tab:gru}
        \centering
         \begin{tabular}{@{}lcccc}
         \hline
             Model & Horizon & MAE & MSE & RMSE \\ \hline
              GRU & 3 & 1.2119 & 2.4764 & 1.5736 \\
              Bi-GRU & 3 & 1.1049 & 2.1991 & 1.4829 \\
              SENT-Bi-GRU & 3 & \textbf{1.0411}  & \textbf{2.0097} & \textbf{1.4176} \\
              USD-Bi-GRU & 3 & 1.0874 & 2.1378 & 1.4621 \\
              TENI-Bi-GRU & 3 & 1.0960 & 2.1537 & 1.4675 \\

              SENT-GRU-1 & 1 & \textbf{0.0647} & \textbf{0.0069} & \textbf{0.0833} \\
              TENI-GRU-1 & 1 &0.3104 & 0.1530 & 0.3912 \\ \hline
              
         \end{tabular} 
        \vspace{3mm}
        \hrule
        \centering
        \end{table}
    
    \begin{table}[!hbt]
    \hrule
    \setlength{\tabcolsep}{0.96pc}
    \caption{LSTM-based models}
    \label{tab:lstm}
    \centering
         \begin{tabular}{@{}lcccc}
         \hline
             Model & Horizon & MAE & MSE & RMSE \\ \hline
              SENT-Bi-LSTM & 3 & 1.0455  & 2.0231 & 1.4223 \\
              USD-Bi-LSTM & 3 & 1.1690 & 2.3917 & 1.5465 \\
              TENI-Bi-LSTM & 3 & 1.1570 & 2.3038 & 1.5178 \\
              SENT-Bi-CNN-LSTM & 3 & 1.3955 & 3.3865 & 1.8402 \\
              SENT-Bi-CNN-LSTM-att & 3 & 1.4524 & 3.2913 & 1.8141 \\
            \hline
        \end{tabular} 
        \vspace{3mm}
        \hrule
        \centering
    \end{table}

    \subsection{Transformers and TimesNet Experiments}
    \begin{table}[!hbt]
    \hrule
    \setlength{\tabcolsep}{1pc}
    \caption{Transformer-based models}
    \label{tab:transformer-result-table}
    \begin{tabular}{l|c|ccc}
    \hline
    &\begin{tabular}[c]{@{}c@{}}prediction \\ Length\end{tabular} & \textbf{MAE} & \textbf{MSE} & \textbf{RMSE} \\ \hline
         & 1 & 1.3854 & 3.2847 & 1.8123 \\ \cline{2-5} 
        \multirow{-2}{*}{SENT-Autoformer} & 3 & \multicolumn{1}{l}{{2.8073}} & 5.7048 & 2.3884 \\ \hline
         & 1 & 3.7948 & 18.6935 & 4.3236 \\ \cline{2-5} 
        \multirow{-2}{*}{SENT-Informer} & 3 & 5.0463 & 31.8270 & 5.6415 \\ \hline
         & 1 & 1.4099 & 3.3947 & 1.8424 \\ \cline{2-5} 
        \multirow{-2}{*}{SENT-TimesNet} & 3 & { 3.4176} & 7.9913 & 2.8269 \\ \hline
         & 1 & 3.4512 & 15.251 & 3.9052 \\ \cline{2-5} 
        \multirow{-2}{*}{SENT-Transformer} & 3 & 6.5671 & 30.4323 & 5.5165 \\ \hline
         & 1 & 1.5357 & 3.8618 & 1.9651 \\ \cline{2-5} 
        \multirow{-2}{*}{Autoformer} & 3 & 5.383 & 36.9292 & 6.0769 \\ \hline
    \end{tabular} 
    \vspace{3mm}
    \hrule
    \centering
    \end{table}

         We examined a base Transformer model along with two well-known transformer-based architectures specifically designed for time series tasks, namely Autoformer \cite{abs-2106-13008} and Informer \cite{2012-07436}. Additionally, we examined another advanced time series model named TimesNet, which is based on a CNN architecture using a temporal 2D-variation modelling approach \cite{wu2023timesnet}. Similar to the experiments presented in the previous section, we conduct experiments incorporating the three external variables in various combinations. Among the models mentioned above, the best results were obtained regarding 3-day ahead forecasting by incorporating the sentiment score with Autoformer (MAE 2.8073, MSE 5.7048, RMSE 2.3884) as shown in Table (\ref{tab:transformer-result-table}).
         These transformer-based models exhibit inferior performance compared to sequence models. This disparity can be attributed to the self-attention mechanism applied to transformer-based models, which is somewhat “anti-order”  and can lead to temporal information loss. This loss is usually not a significant concern in semantic-rich applications like natural language processing (NLP), where the meaning of a sentence remains largely preserved even if the word order is altered. However, when dealing with time series data, the primary focus is modelling the temporal relation within a continuous sequence of points.  Additionally, it's essential to acknowledge the relatively higher training time and complexities associated with Transformers regarding the difficulty in fine-tuning the hyperparameters compared to more traditional models like GRU or LSTM.

    \subsection{External variables}  
        The impact of three external time series variables on prediction accuracy has been investigated: USDX, TENI, and the cumulative sentimental score (SENT). The initial part of the models' names indicates the integrated variables; for example, "SENT-Bi-GRU" signifies a Bi-GRU model where the sentiment index serves as an input. 
        The results shown in Tables (\ref{tab:gru} and \ref{tab:lstm}) suggest that the cumulative sentimental score (SENT) is the most effective external time series variable for improving the prediction accuracy of Brent crude oil price. SENT is more effective because it captures the overall sentiment of the market, which can be a valuable predictor of future prices. USDX and TENI, on the other hand, may not be as effective predictors because they only capture specific aspects of the market, such as the strength of the US dollar or the price of oil.
       
    \subsection{SENT-Bi-GRU}
        Among all the previous experiments conducted, the SENT-BI-GRU network, which is a Bi-GRU network integrated with sentiment index,  demonstrated the highest accuracy performance. SENT-Bi-GRU is a simple and effective bidirectional GRU architecture incorporating a cumulative sentiment index. The architecture of this network consists of a bidirectional GRU network followed by two fully connected layers, all connected through ReLU activation functions. The best performance has been achieved with a window size of 5 time steps, a batch size of 16, and the AdamW optimiser. 
        
    \subsection{ERS-Bi-GRU}  
    \label{sec:ers-bigru}
\begin{figure}[!hbt]
        \centering
        \includegraphics[scale=0.25]{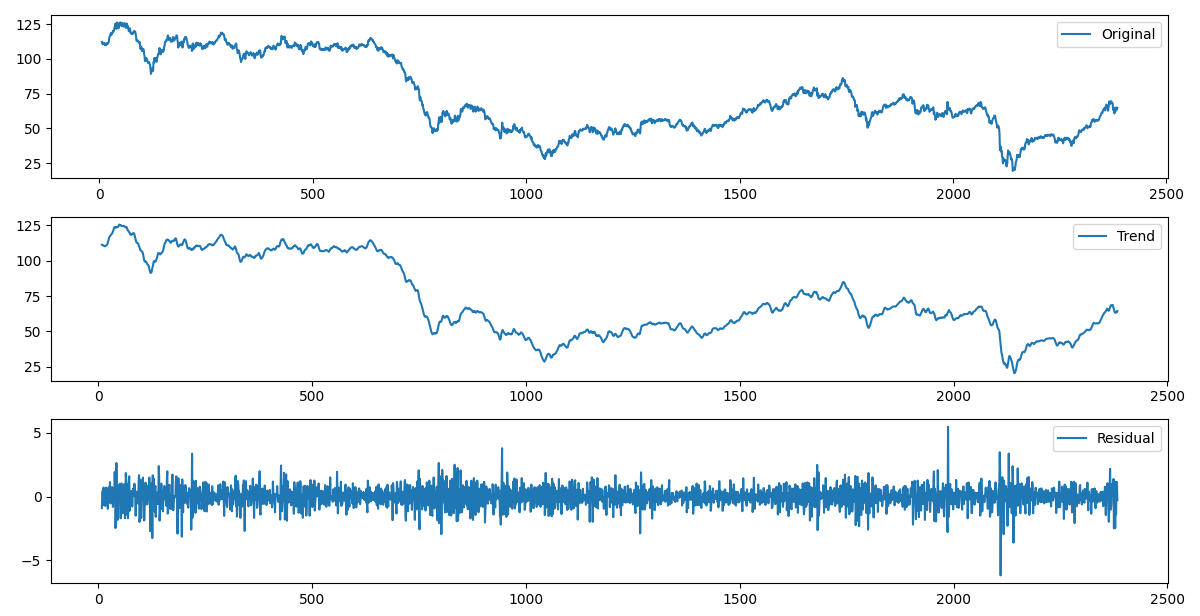}
        \caption{Crude Brent oil price components}
        \label{fig:residual}
    \end{figure}

    To further refine the predictive capabilities, we incorporated a residual analysis into the previous model (SENT-Bi-GRU).
    The main idea is to ensemble the forecasts of two Bi-GRU networks. One network incorporates a sentimental index as an input feature, while the other incorporates a residual component as an input feature (Figure \ref{fig:residual} depicts the residual component). The forecasts from these two networks are then fused using a weighted averaging method, resulting in a more robust and accurate prediction.
   The objective is to enhance capturing fluctuations and irregularities in the oil market. We assume that we can model the price fluctuations with the help of the sentiment index and the residual component.
   
   The proposed models, denoted as Ensemble Sentiment-Residual BiGRU (ESR-BiGRU), involve the following steps:
    \begin{enumerate}
        \item Train a Bi-GRU model to future crude oil prices. The model takes the historical crude oil prices and sentiment index as inputs. Let $P_t$ represent the crude oil price at time $t$ and $S_t$ denote the sentiment index simultaneously. The forecast from Bi-GRU can be denoted as $forecast_1 = F_{\text{Bi-GRU}}(P_t, S_t)$
        
        \item Train a Bi-GRU model: This model takes sentiment index $S$ and the residual component $R$ (calculated by removing the trend from the oil prices) as inputs and learns to predict the unpredictable variations in oil prices. Let $R_t$ represent the residual at time $t$ and $S_t$ denote the sentiment index simultaneously. The forecast from Bi-GRU can be denoted as $forecast_2 = F_{\text{Bi-GRU}}(R_t, S_t)$
        
        \item Train a Bi-GRU model: This model takes historical crude oil prices $P$ and the residual component $R$ as inputs and learns to predict future crude oil prices. The forecast from this Bi-GRU is denoted as $forecast_3 = F_{\text{Bi-GRU}}(P_t, R_t)$
        
        \item Calculate the final forecast: The last forecast is obtained by combining the three forecasts as follows:
        $Forcat_{final} = ((forecast_3 - forecast_2 * W_1) * W_2) + (forecast_1 * W_3) $.
    \end{enumerate}

    \begin{figure}
        \centering
        \includegraphics[width=0.40\textwidth]{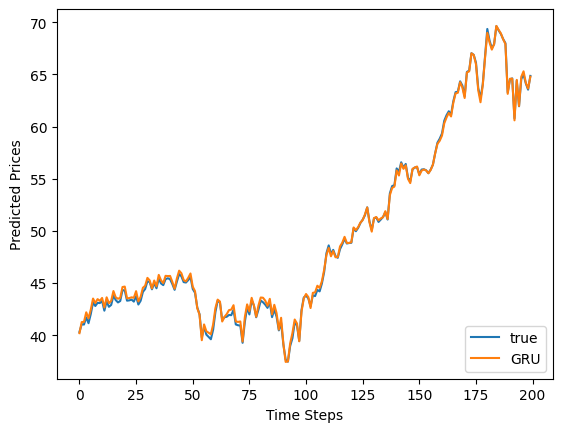}
        \includegraphics[width=0.40\textwidth]{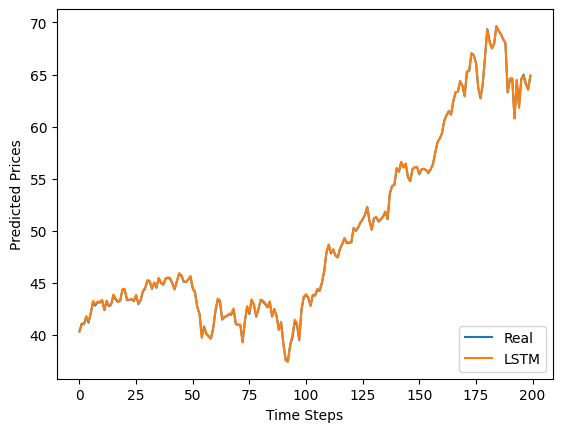}

        \caption{One-ahead day forecasts using SENT-GRU, SENT-LSTM models}
        \label{fig:1day}
    \end{figure}
       
      This model outperforms all the other models examined in this paper and benchmark models, achieving MAE 1.04475, MSE 1.9946 and RMSE 1.4123 
 
     \begin{table}[!hbt]
         \hrule
         \setlength{\tabcolsep}{0.80pc}
         \caption{ERS-Bi-GRU model results versus other created benchmark models. The values are calculated using the actual, not scaled prices.}
         \label{tab:main-comp-3}
         \centering
         \begin{tabular}{lcccc}
         \hline
              & Horizon & MAE & MSE & RMSE \\ \hline
              ERS-Bi-GRU &3 & \textbf{1.04475}& \textbf{1.9946} & \textbf{1.4123} \\
              Bi-GRU & 3 & 1.1049 & 2.1991 & 1.4829 \\
              SENT-Bi-GRU & 3 & 1.0411  & 2.0097 & 1.4176 \\
              SENT-Bi-LSTM & 3 & 1.0455  & 2.0231 & 1.4223 \\            
              SENT-Bi-CNN-LSTM & 3 & 1.3955 & 3.3865 & 1.8402 \\
              SENT-Bi-CNN-LSTM-att & 3 & 1.4524 & 3.2913 & 1.8141 \\

              SENT-Autoformer & 3 & 2.8073 & 5.7048 & 2.3884 \\

              SENT-TimesNet & 3 & 3.4176  & 7.9913 & 2.8269 \\
              \hline
      \end{tabular} 
    \vspace{3mm}
    \hrule
    \centering
    \end{table}
    
    \subsection{Discussion and benchmark comparison}
    \begin{figure}[!hbt]
        \centering
        \includegraphics[width=0.24\textwidth]{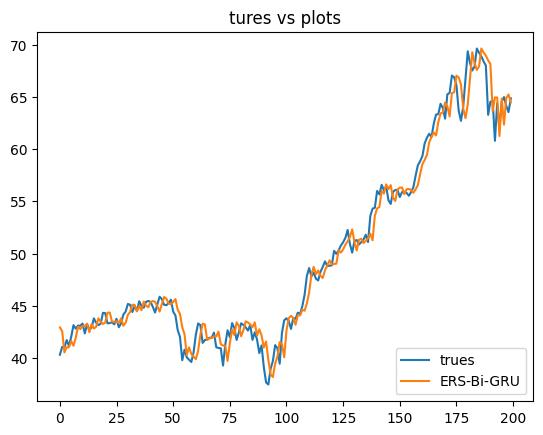}
        \includegraphics[width=0.24\textwidth]{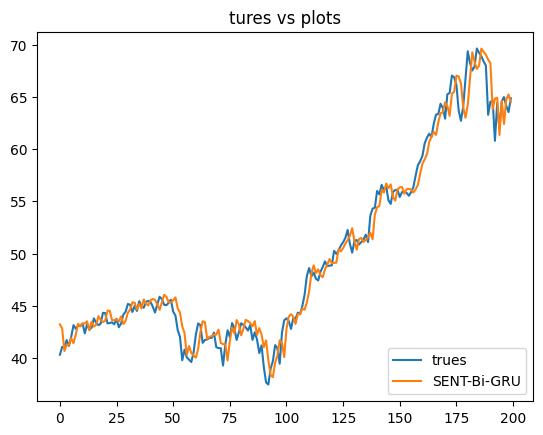}
        \includegraphics[width=0.24\textwidth]{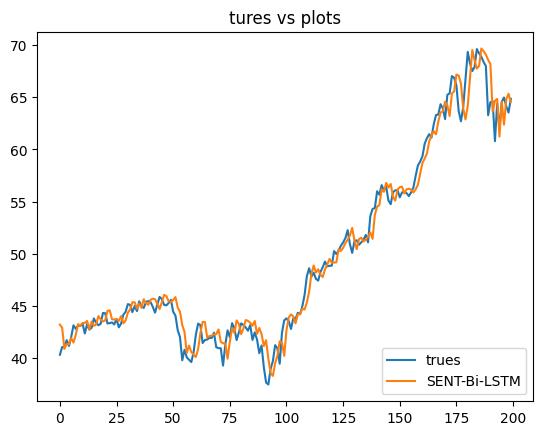}
        \includegraphics[width=0.24\textwidth]{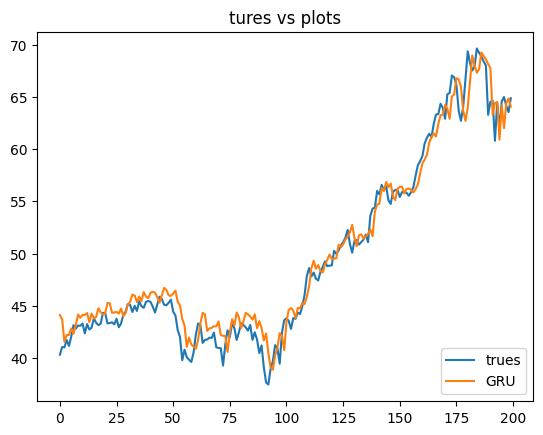}
        \includegraphics[width=0.26\textwidth]{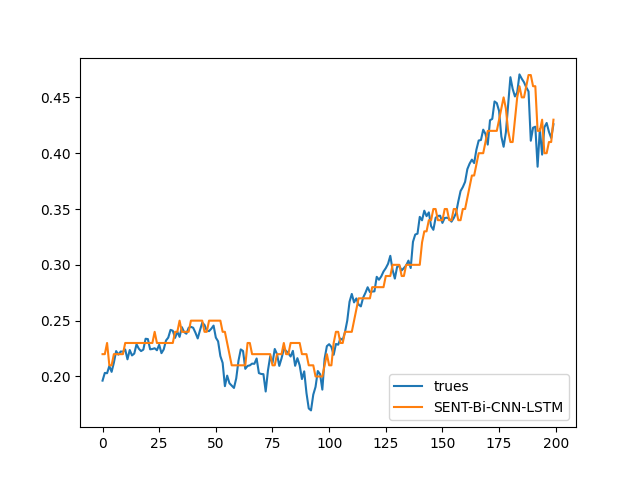}
        \includegraphics[width=0.26\textwidth]{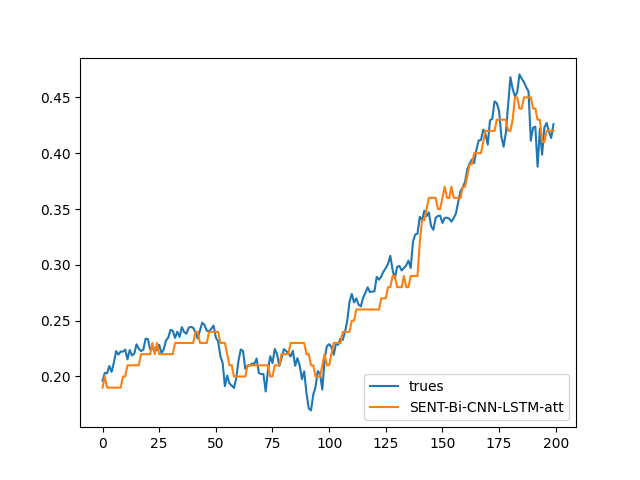}
        \includegraphics[width=0.22\textwidth]{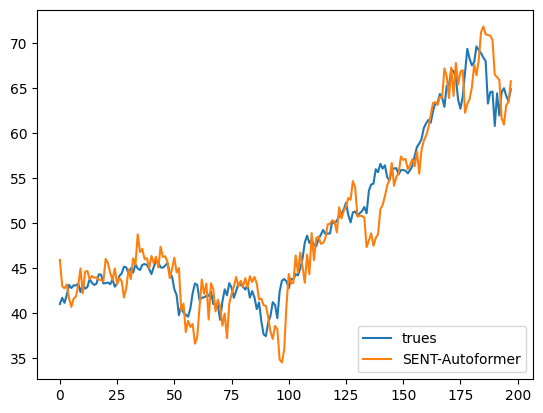}
        \includegraphics[width=0.22\textwidth]{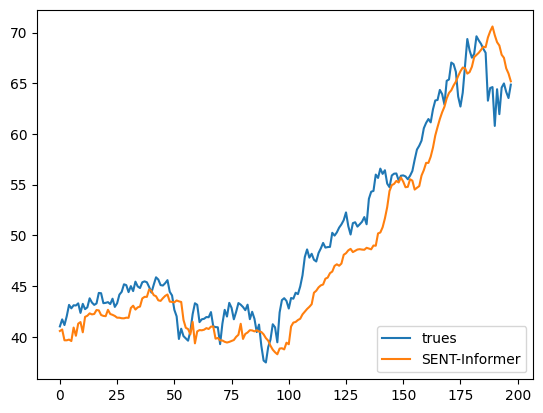}
        \caption{3-ahead day forecasts, ERS-Bi-GRU model versus other benchmark models}
        \label{fig:3day}
    \end{figure}
    
         In general, the significance of the experiment results is as follows: First,  simple models (GRU and LSTM) outperform transformer and TimesNet models for Brent crude oil price forecasting; Second, incorporating crude oil-relevant sentimental index is demonstrated to be effective and promising to help in estimating the fluctuation level of the price.
         However, among the various models considered, our findings reveal that Bidirectional (LSTM/GRU) models yield the best performance. While Bi-GRU outperforms slightly the Bi-LSTM. We obtained the best results of forecasting with the ESR-Bi-GRU model (MAE 1.04475, MSE 1.9946, and RMSE 1.4123)  for a 3-days ahead forecasting as shown in Table (\ref{tab:main-comp-3}).  It is worth mentioning that MAE, MSE, and RMSE have been calculated using the re-scaled predicted values. Optimal time-step values for different models are among  (5, 10,17, and 22), with 16 of batch size.
        
        Figures (\ref{fig:1day}, \ref{fig:3day}) depict samples of the forecasting results of the proposed model and the other created benchmark models. The proposed model (ERS-Bi-GRU) exhibits a high level of accuracy in forecasting the closing prices of Brent crude oil, demonstrating a solid alignment between its predictions and the observed values. This highlights the model's ability to provide precise forecasts for crude oil prices. 

\section{Conclusion} 
This work follows a comprehensive methodology to achieve accurate multi-step forecasting for crude Brent prices. 
        We started by conducting a comparative analysis of the performance of various architectures of six deep learning networks(GRU, LSTM, Autoformer, Informer, Transformer, TimesNet) for Brent crude price forecasting. In addition, it introduces a sample and efficient model, ERS-Bi-GRU, for multi-step forecasting of Brent crude price. Our experiments go beyond the models by examining the significance of external factors that commonly influence the crude oil market. By incorporating these additional factors, the evaluation aims to provide a holistic understanding of the models' forecasting capabilities in real-world scenarios. The results demonstrated that sequence-based models (GRU and LSTM) outperform transformer-based models for forecasting crude oil prices. The results showed that the proposed ERS-Bi-GRU model outperforms benchmark models in the field and state-of-the-art models (Autoformer and TimesNet) regarding one/multi-step forecasting. The results also demonstrated the importance of extraneous factors to improve the forecasting accuracy. Finally, it is worth mentioning that the nature of the data significantly influences training models for forecasting tasks and is extremely sensitive to hyperparameter configurations. Therefore, efficient algorithms for optimising hyper-parameter configuration are recommended.

        In future work, we seek to enhance this model by integrating optimisation algorithms (such as Particle, Swarm Optimization, Gravity Search, algorithm, and Gray Wolf Optimizer) for hyperparameter tuning. Additionally, we plan to incorporate the proposed model within a GAN architecture and add SDE solvers for modelling volatility and noise. Furthermore, we intend to apply this approach to financial data.

\section*
{Data Availability}
The dataset used in this paper can be accessed through this \href{https://github.com/Med-Rokaimi/Exploratory-data-analysis--Brent-crude-oil} {GitHub repos}
\section*{Funding}
No external funding was received for this research.

%\begin{thebibliography}{999}
\bibliographystyle{elsarticle-num} 
\bibliography{ref.bib}

\section*{Ethics Approval}
Not applicable.

\section*{Corresponding author}
Corresponding author: Mohammed Alruqimi

\end{document}